# A Multidimensional Cascade Neuro-Fuzzy System with Neuron Pool Optimization in Each Cascade

**Yevgeniy V. Bodyanskiy**
Kharkiv National University of Radio Electronics, Kharkiv, Ukraine,
Email: bodya@kture.kharkov.ua

**Oleksii K. Tyshchenko and Daria S. Kopaliani**
Kharkiv National University of Radio Electronics, Kharkiv, Ukraine,
Email: { lehatish, daria.kopaliani }@gmail.com

*Abstract*— A new architecture and learning algorithms for the multidimensional hybrid cascade neural network with neuron pool optimization in each cascade are proposed in this paper. The proposed system differs from the well-known cascade systems in its capability to process multidimensional time series in an online mode, which makes it possible to process non-stationary stochastic and chaotic signals with the required accuracy. Compared to conventional analogs, the proposed system provides computational simplicity and possesses both tracking and filtering capabilities.

## I. INTRODUCTION

Today artificial neural networks (ANNs) and neuro-fuzzy systems (NFSs) are successfully used in a wide range of data processing problems (when data can be presented either in the form of "object-property" tables or in the form of time series, often produced by non-stationary nonlinear stochastic or chaotic systems). The advantages ANNs and NFSs have over other existing approaches derive from their universal approximating capabilities and learning capacities.

Conventionally "learning" is defined as a process of adjusting synaptic weights using an optimization procedure that involves searching for the extremum of a given learning criterion. The learning process quality can be improved by adjusting a network topology along with its synaptic weights [1, 2]. This idea is the foundation of evolving computational intelligence systems [3, 4]. One of the most successful implementations of this approach is cascade-correlation neural networks [5–8] due to their high degree of efficiency and learning simplicity of both synaptic weights and a network topology). Such a network starts off with a simple architecture consisting of a pool (ensemble) of neurons which are trained independently (the first cascade). Each neuron in the pool can have a different activation function and a different learning algorithm. The neurons in the pool do not interact with each other while they are trained. After all the neurons in the pool of the first cascade have had their weights adjusted, the best neuron with respect to a learning criterion forms the first cascade and its synaptic weights can no longer be adjusted. Then the second cascade is formed usually out of similar neurons in the training pool. The only difference is that neurons which are trained in the pool of the second cascade have an additional input (and therefore an additional synaptic weight) which is an output of the first cascade. Similar to the first cascade, the second cascade will eliminate all but one neuron showing the best performance whose synaptic weights will thereafter be fixed.

Neurons of the third cascade have two additional inputs, namely the outputs of the first and second cascades. The evolving network continues to add new cascades to its architecture until it reaches the desired quality of problem solving over the given training set.

Authors of the most popular cascade neural network, CasCorLA, S. E. Fahlman and C. Lebiere, used elementary Rosenblatt perceptrons with traditional sigmoidal activation functions and adjusted synaptic weights using the Quickprop-algorithm (a modification of the $\delta$-learning rule). Since the outgoing signal of such neurons is non-linearly dependent on its synaptic weights, the learning rate cannot be increased for such neurons. In order to avoid multi-epoch learning [9–16], different types of neurons (with outputs that depend linearly on synaptic weights) should be used as network nodes. This would allow the use of optimal learning algorithms in terms of speed and process data as it is an input to the network. However, if the network is learning in an online mode, it is impossible to determine the best neuron in the pool. While working with non-stationary objects, one neuron of the training pool can be identified as the best for one part of the training set, but not for the others. Thus we suggest that all neurons retain in the training pool and a certain optimization procedure (generated according to a general network quality criterion) is used to determine an output of the cascade.

It should be noticed that the well-known cascade neural networks implement non-linear mapping $R^n \to R^1$, i.e. they are a systems with a single output. At the same time, many problems (which are solved with the help of ANNs and NFSs) require the multidimensional mapping $R^n \to R^g$ implementation, which leads to the fact that $g$ times more neurons should be trained in each cascade comparing to a conventional neural network, which makes such a system too cumbersome. Therefore, it seems appropriate to use as the cascade network nodes specialized multidimensional neuron structures with multiple outputs instead of traditional neurons like elementary Rosenblatt perceptrons.

The remainder of this paper is organized as follows: Section 2 gives an optimized multidimensional cascade neural network architecture. Section 3 describes training neo-fuzzy neurons in the network. Section 4 describes output signals' optimization of the multidimensional neo-fuzzy neuron pool. Section 5 presents experiments and evaluation. Conclusions and future work are given in the final section.

## II. AN OPTIMIZED MULTIDIMENSIONAL CASCADE NEURAL NETWORK ARCHITECTURE

An input of the network (the so-called "receptive layer") is a vector signal $x(k) = (x_1(k), x_2(k), \ldots, x_n(k))^T$, where $k = 1, 2, \ldots$ is either the quantity of samples in the "object-property" table or the current discrete time. These signals are sent to the inputs of each neuron $MN_j^{[m]}$ in the network ($j = 1, 2, \ldots, q$ is the quantity of neurons in a training pool, $m = 1, 2, \ldots$ is a cascade number), that produces vector outputs $\hat{y}_d^{[m]j}(k) = \left(\hat{y}_1^{[m]j}(k), \hat{y}_2^{[m]j}(k), \ldots, \hat{y}_g^{[m]j}(k)\right)^T$, $d = 1, 2, \ldots, g$. These outputs are later combined with a generalizing neuron $GMN^{[m]}$ which generates an optimal vector output $\hat{y}^{*[m]}(k)$ of the $m$-th cascade. While the input of the neurons in the first cascade is $x(k)$, neurons in the second cascade have $g$ additional inputs for the generated signal $\hat{y}^{*[1]}(k)$, neurons in the third cascade have $2g$ additional inputs $\hat{y}^{*[1]}(k), \hat{y}^{*[2]}(k)$, neurons in the $m$-th cascade have $(m-1)g$ additional inputs $\hat{y}^{*[1]}(k), \hat{y}^{*[2]}(k), \ldots, \hat{y}^{*[m-1]}(k)$. New cascades become a part of the network during a training process when it becomes clear that the current cascades do not provide the desired quality of information processing.

## III. TRAINING NEO-FUZZY NEURONS IN THE MULTIDIMENSIONAL CASCADE NEURO-FUZZY NETWORKS

The low learning rate of the Rosenblatt perceptrons (which are widely used in traditional cascade ANNs) along with the difficulty in interpreting results (inherent to all ANN in general) encourages us to search for alternative approaches to the synthesis of evolving systems in general and cascade neural networks in particular. High interpretability and transparency along with good approximation capabilities and ability to learn are the main features of neuro-fuzzy systems [17, 18], which are the foundation of hybrid computational intelligence systems. In [10,11,13] hybrid cascade systems were introduced which used neo-fuzzy neurons [19-21] as network nodes, allowing one to significantly increase the speed of synaptic weight adjustment. A neo-fuzzy neuron (NFN) is a nonlinear system having the following mapping

$$\hat{y} = \sum_{i=1}^{n} f_i(x_i) \qquad (1)$$

where $x_i$ is the $i$-th input $(i = 1, 2, \ldots, n)$, $\hat{y}$ is the neo-fuzzy neuron's output. Structural units of the neo-fuzzy neuron are nonlinear synapses $NS_i$ which transform the $i$-th input signal in the following way

$$f_i(x_i) = \sum_{l=1}^{h} w_{li} \mu_{li}(x_i) \qquad (2)$$

where $w_{li}$ is the $l$-th synaptic weight of the $i$-th nonlinear synapse, $l = 1, 2, \ldots h$ is the total quantity of synaptic weights and therefore membership functions $\mu_{li}(x_i)$ in this nonlinear synapse. In this way $NS_i$ implements the fuzzy inference [18]

$$IF\ x_i\ IS\ X_{li}\ THEN\ THE\ OUTPUT\ IS\ w_{li} \qquad (3)$$

where $X_{li}$ is a fuzzy set with a membership function $\mu_{li}$, $w_{li}$ is a singleton (a synaptic weight in a consequent). It can be seen that, in fact, a nonlinear synapse implements Takagi-Sugeno zero-order fuzzy inference.

The $j$-th neo-fuzzy neuron $(j = 1, 2, \ldots, q)$ of the $d$-th output $(d = 1, 2, \ldots, g)$ of the first cascade (according to the network topology) can be written

$$\begin{cases} \hat{y}_d^{[1]j}(k) = \sum_{i=1}^{n} f_{di}^{[1]j}(x_i(k)) = \sum_{i=1}^{n} \sum_{l=1}^{h} w_{dli}^{[1]j} \mu_{dli}^{[1]j}(x_i(k)), \\ IF\ x_i(k)\ IS\ X_{li}^{j}\ THEN\ THE\ OUTPUT\ IS\ w_{dli}^{[1]j}. \end{cases} \quad (4)$$

Authors of the neo-fuzzy neuron [19–21] used traditional triangular constructions meeting the conditions of Ruspini partitioning (unity partitioning) as membership functions [22]:

$$\mu_{dli}^{[1]j}(x_i) = \begin{cases} \dfrac{x_i - c_{d,l-1,i}^{[1]j}}{c_{dli}^{[1]j} - c_{d,l-1,i}^{[1]j}}, & if\ x_i \in \left[c_{d,l-1,i}^{[1]j}, c_{dli}^{[1]j}\right], \\ \dfrac{c_{d,l+1,i}^{[1]j} - x_i}{c_{d,l+1,i}^{[1]j} - c_{dli}^{[1]j}}, & if\ x_i \in \left[c_{dli}^{[1]j}, c_{d,l+1,i}^{[1]j}\right], \\ 0, otherwise \end{cases} \quad (5)$$

where $c_{dli}^{[1]j}$ are relatively arbitrarily chosen (usually evenly distributed) center parameters of membership functions over the interval $[0,1]$ where, naturally, $0 \le x_i \le 1$. This choice of membership functions ensures that the input signal $x_i$ only activates two neighboring membership functions, and their sum is always equal to 1 which means that

$$\mu_{dli}^{[1]j}(x_i) + \mu_{d,l+1,i}^{[1]j}(x_i) = 1 \quad (6)$$

and

$$f_{di}^{[1]j}(x_i) = w_{dli}^{[1]j} \mu_{dli}^{[1]j}(x_i) + w_{d,l+1,i}^{[1]j} \mu_{d,l+1,i}^{[1]j}(x_i). \quad (7)$$

It is clear that other constructions such as polynomial and harmonic functions, wavelets, orthogonal functions, etc. can be used as membership functions in nonlinear synapses. It is still unclear which of the functions provide the best results, which is why the idea of using not one neuron, but a pool of neurons with different membership and activation functions seems promising.

Similar to (4) we can determine outputs for the remaining cascades: outputs of the neurons in the second cascade:

$$\hat{y}_d^{[2]j} = \sum_{i=1}^{n} \sum_{l=1}^{h} w_{dli}^{[2]j} \mu_{dli}^{[2]j}(x_i) + \\ + \sum_{d=1}^{g} \sum_{l=1}^{h} w_{dl,n+1}^{[2]j} \mu_{dl,n+1}^{[2]j}\left(\hat{y}_d^{*[1]}\right) \forall\ d = 1, 2, \ldots, g; \quad (8)$$

$\vdots$

outputs of the neurons in the $m$-th cascade:

$$\hat{y}_d^{[m]j} = \sum_{i=1}^{n} \sum_{l=1}^{h} w_{dli}^{[m]j} \mu_{dli}^{[m]j}(x_i) + \\ + \sum_{d=1}^{g} \sum_{p=n+1}^{n+m-1} \sum_{l=1}^{h} w_{dlp}^{[m]j} \mu_{dlp}^{[m]j}\left(\hat{y}_d^{*[p-n]}\right) \forall\ d = 1, 2, \ldots, g. \quad (9)$$

Thus, the cascade network formed with neo-fuzzy neurons consisting of $m$ cascades contains $gh\left(n + \sum_{p=1}^{m-1} p\right)$ parameters. Introducing the vector of membership functions for the $j$-th NFN of the $d$-th output in the $m$-th cascade

$$\mu_d^{[m]j}(k) = \left( \mu_{d11}^{[m]j}(x_1(k)), \ldots, \mu_{dh1}^{[m]j}(x_1(k)), \mu_{d12}^{[m]j}(x_2(k)), \ldots, \mu_{dh2}^{[m]j}(x_2(k)), \ldots, \mu_{dli}^{[m]j}(x_i(k)), \ldots, \right.$$
$$\left. \mu_{dhn}^{[m]j}(x_n(k)), \mu_{d1,n+1}^{[m]j}(\hat{y}_1^{*[1]}(k)), \ldots, \mu_{dh,n+h}^{[m]j}(\hat{y}_g^{*[1]}(k)), \ldots, \mu_{dh,g(n+m-1)}^{[m]j}(\hat{y}_g^{*[m-1]}(k)) \right)^T$$

and a corresponding synaptic weights' vector

$$w_d^{[m]j} = \left( w_{d11}^{[m]j}, \ldots, w_{dh1}^{[m]j}, w_{d12}^{[m]j}, \ldots, w_{dh2}^{[m]j}, \ldots, w_{dli}^{[m]j}, \ldots, w_{dhn}^{[m]j}, w_{d1,n+1}^{[m]j}, \ldots, w_{dh,n+h}^{[m]j}, \ldots, w_{dh,g(n+m-1)}^{[m]j} \right)^T,$$

the output signal can be finally written down in a compact form

$$\hat{y}_d^{[m]j}(k) = \left( w_d^{[m]j} \right)^T \mu_d^{[m]j}(k). \qquad (10)$$

Since this signal is linearly dependent on the synaptic weights, any adaptive identification algorithm [23-25] can be used for training the network neo-fuzzy-neurons, for example, the exponentially-weighted least-squares method in a recurrent form

$$\begin{cases} w_d^{[m]j}(k+1) = w_d^{[m]j}(k) + \left( \alpha + \left( \mu_d^{[m]j}(k+1) \right)^T \times \right. \\ \left. \times P_d^{[m]j}(k) \mu_d^{[m]j}(k+1) \right)^{-1} \left( P_d^{[m]j}(k) \left( y^d(k+1) - \right. \right. \\ \left. \left. - \left( w_d^{[m]j}(k) \right)^T \mu_d^{[m]j}(k+1) \right) \mu_d^{[m]j}(k+1), \right. \\ P_d^{[m]j}(k+1) = \frac{1}{\alpha} \left( P_d^{[m]j}(k) - \left( \alpha + \left( \mu_d^{[m]j}(k+1) \right)^T \times \right. \right. \\ \left. \left. \times P_d^{[m]j}(k) \mu_d^{[m]j}(k+1) \right)^{-1} \left( P_d^{[m]j}(k) \mu_d^{[m]j}(k+1) \times \right. \right. \\ \left. \left. \left( \mu_d^{[m]j}(k+1) \right)^T P_d^{[m]j}(k) \right) \right. \end{cases} \qquad (11)$$

(here $y^d(k+1), d = 1, 2, \ldots, g$ – an external learning signal, $0 < \alpha \leq 1$ – a forgetting factor) or the gradient learning algorithm with both tracking and filtering properties [26]

$$\begin{cases} w_d^{[m]j}(k+1) = w_d^{[m]j}(k) + \left( r_d^{[m]j}(k+1) \right)^{-1} \left( y^d(k+1) - \right. \\ \left. - \left( w_d^{[m]j}(k) \right)^T \mu_d^{[m]j}(k+1) \right) \mu_d^{[m]j}(k+1), \qquad (12) \\ r_d^{[m]j}(k+1) = \alpha r_d^{[m]j}(k) + \left\| \mu_d^{[m]j}(k+1) \right\|^2, \ 0 \leq \alpha \leq 1. \end{cases}$$

An architecture of a typical neo-fuzzy neuron which was discussed earlier as a part of the multidimensional neuron $MN_g^{[1]}$ of the cascade system is redundant, since a vector of input signals $x(k)$ (the first cascade) is fed to one-type nonlinear synapses $NS_{di}^{[1]j}$ of the neo-fuzzy neurons, each neuron of which generates at its output a signal $\hat{y}_d^{[1]j}(k), d = 1, 2, \ldots, g$. As a result, the output vector components $\hat{y}^{[1]j}(k) = \left( \hat{y}_1^{[1]j}(k), \hat{y}_2^{[1]j}(k), \ldots, \hat{y}_g^{[1]j}(k) \right)^T$ are calculated independently. This can be avoided by considering a multidimensional neo-fuzzy neuron [27], which is a modification of the system proposed in [28]. Structure nodes are composite nonlinear synapses $MNS_i^{[1]j}$, each synapse contains $h$ membership functions $\mu_{li}^{[1]j}$ and $gh$ adjustable synaptic weights $w_{dli}^{[1]j}$. Thus, a multidimensional neo-fuzzy neuron of the first cascade contains $ghn$ synaptic weights, but only $hn$ membership functions that makes it $g$ times less than if the cascade would have been formed of conventional neo-fuzzy neurons.

Taking into consideration a $(hn \times 1)$ – membership functions' vector

$$\mu^{[1]j}(k) = \left( \mu_{11}^{[1]j}(x_1(k)), \mu_{21}^{[1]j}(x_1(k)), \ldots, \mu_{h1}^{[1]j}(x_1(k)), \right.$$
$$\left. \ldots, \mu_{li}^{[1]j}(x_i(k)), \ldots, \mu_{hn}^{[1]j}(x_n(k)) \right)^T$$

and a $(g \times hn)$ – synaptic weights' matrix

$$W^{[1]j} = \begin{pmatrix} w_{111}^{[1]j} & w_{112}^{[1]j} & \cdots & w_{1li}^{[1]j} & \cdots & w_{1hn}^{[1]j} \\ w_{211}^{[1]j} & w_{212}^{[1]j} & \cdots & w_{2li}^{[1]j} & \cdots & w_{2hn}^{[1]j} \\ \vdots & \vdots & \vdots & \vdots & \vdots & \vdots \\ w_{g11}^{[1]j} & w_{g12}^{[1]j} & \cdots & w_{gli}^{[1]j} & \cdots & w_{ghn}^{[1]j} \end{pmatrix},$$

the output signal $MN_j^{[1]}$ can be written down at the $k$-th time moment in the form of

$$\hat{y}^{[1]j}(k) = W^{[1]j} \mu^{[1]j}(k). \qquad (13)$$

Multidimensional neo-fuzzy-neuron learning can be implemented with the help of a matrix modification of the exponentially-weighted recurrent least squares method (11) in the form of

$$\begin{cases} W^{[1]j}(k+1) = W^{[1]j}(k) + \left(\alpha + \left(\mu^{[1]j}(k+1)\right)^T P^{[1]j}(k) \times \right. \\ \left. \times \mu^{[1]j}(k+1)\right)^{-1} \left(y(k+1) - W^{[1]j}(k) \mu^{[1]j}(k+1)\right) \times \\ \times \left(\mu^{[1]j}(k+1)\right)^T P^{[1]j}(k), \\ P^{[1]j}(k+1) = \frac{1}{\alpha}\left(P^{[1]j}(k) - \left(\alpha + \left(\mu^{[1]j}(k+1)\right)^T \times \right.\right. \\ \left.\left. \times P^{[1]j}(k) \mu^{[1]j}(k+1)\right)^{-1} \left(P^{[1]j}(k) \mu^{[1]j}(k+1) \times \right.\right. \\ \left.\left. \times \left(\mu^{[1]j}(k+1)\right)^T P^{[1]j}(k)\right)\right), 0 < \alpha \leq 1 \end{cases} \qquad (14)$$

or in the form of a multidimensional version of the algorithm (12) [29]:

$$\begin{cases} W^{[1]j}(k+1) = W^{[1]j}(k) + \left(r^{[1]j}(k+1)\right)^{-1} \left(y(k+1) - \right. \\ \left. - W^{[1]j}(k) \mu^{[1]j}(k+1)\right) \left(\mu^{[1]j}(k+1)\right)^T, \\ r^{[1]j}(k+1) = \alpha r^{[1]j}(k) + \left\|\mu^{[1]j}(k+1)\right\|^2, \ 0 \leq \alpha \leq 1 \end{cases} \qquad (15)$$

here $y(k+1) = \left(y^1(k+1), y^2(k+1), \ldots, y^g(k+1)\right)^T$.

The remaining cascades are trained in a similar manner, while the $m$-th cascade membership functions' vector $\mu^{[m]j}(k+1)$ increases its dimensionality by $(m-1)g$ components which are formed by the previous cascades' outputs.

IV. OUTPUT SIGNALS' OPTIMIZATION OF THE MULTIDIMENSIONAL NEO-FUZZY NEURON POOL

The output signals of all the neurons $MN_d^{[m]}$ of the pool in each cascade are joined by a neuron generalizer $GMN^{[m]}$, the output signals $\hat{y}^{*[m]}(k) = \left(\hat{y}_1^{*[m]}(k), \hat{y}_2^{*[m]}(k), \ldots, \hat{y}_g^{*[m]}(k)\right)^T$ of which should surpass the accuracy of any signal $\hat{y}_j^{[m]}(k)$. This problem can be solved by using undetermined Lagrange multipliers and adaptive multidimensional generalized prediction [30, 31].

Let's introduce a neuron output signal $GMN^{[m]}$ in the form of

$$\hat{y}^{*[m]}(k) = \sum_{j=1}^{q} c_j^{[m]} \hat{y}_j^{[m]}(k) = \hat{y}^{[m]}(k) c^{[m]} \qquad (16)$$

where $\hat{y}^{[m]}(k) = \left(\hat{y}_1^{[m]}(k), \hat{y}_2^{[m]}(k), \ldots, \hat{y}_q^{[m]}(k)\right) - (g \times q)$-a matrix, $c^{[m]}(k) - (q \times 1)$-a generalization coefficients' vector meeting the unbiasedness conditions

$$\sum_{j=1}^{q} c_j^{[m]} = E^T c^{[m]} = 1, \qquad (17)$$

$E = (1,1,\ldots,1)^T - (q \times 1)$ – a vector that consists of ones.

Entering a training criterion

$$E^{[m]}(k) = \sum_{\tau=1}^{k} \left\| y(\tau) - \hat{y}^{[m]}(\tau) c^{[m]} \right\|^2 = Tr\left( \left( Y(k) - \hat{Y}^{[m]}(k) I \otimes c^{[m]} \right)^T \left( Y(k) - \hat{Y}^{[m]}(k) I \otimes c^{[m]} \right) \right) \quad (18)$$

(here $Y(k) = \left( y^T(1), y^T(2), \ldots, y^T(k) \right)^T - (k \times s)$ – an observations' matrix,

$$\hat{Y}^{[m]}(k) = \begin{pmatrix} \hat{y}_1^{[m]T}(1) & \hat{y}_2^{[m]T}(1) & \cdots & \hat{y}_q^{[m]T}(1) \\ \hat{y}_1^{[m]T}(2) & \hat{y}_2^{[m]T}(2) & \cdots & \hat{y}_q^{[m]T}(2) \\ \vdots & \vdots & \vdots & \vdots \\ \hat{y}_1^{[m]T}(k) & \hat{y}_2^{[m]T}(k) & \cdots & \hat{y}_q^{[m]T}(k) \end{pmatrix} - \text{a } (k \times gq) - \text{matrix, } I - \text{an identity } (g \times g) - \text{matrix, } \otimes - \text{a}$$

tensor product symbol).

Let's write down the Lagrange function taking into consideration the constraint (6)

$$L^{[m]}(k) = E^{[m]}(k) + \lambda \left( E^T c^{[m]} - 1 \right) = \sum_{\tau=1}^{k} \left\| y(\tau) - \hat{y}^{[m]}(\tau) c^{[m]} \right\|^2 + \lambda \left( E^T c^{[m]} - 1 \right) =$$

$$= Tr\left( \left( Y(k) - \hat{Y}^{[m]}(k) I \otimes c^{[m]} \right)^T \left( Y(k) - \hat{Y}^{[m]}(k) I \otimes c^{[m]} \right) \right) + \lambda \left( E^T c^{[m]} - 1 \right) = \quad (19)$$

$$= Tr\left( V^{[m]T}(k) V^{[m]}(k) \right) + \lambda \left( E^T c^{[m]} - 1 \right)$$

where $V^{[m]}(k) = Y(k) - \hat{Y}^{[m]}(k) I \otimes c^{[m]} - (k \times g)$ – an innovations' matrix.

The Karush-Kuhn-Tucker equations' system solution

$$\begin{cases} \nabla_{c^{[m]}} L^{[m]}(k) = \vec{0}, \\ \dfrac{\partial L^{[m]}(k)}{\partial \lambda} = 0 \end{cases} \qquad (20)$$

leads to an obvious result

$$\begin{cases} c^{[m]} = \left( R^{[m]}(k) \right)^{-1} E \left( E^T \left( R^{[m]}(k) \right)^{-1} E \right)^{-1}, \\ \lambda = -2 E^T \left( R^{[m]}(k) \right)^{-1} E \end{cases} \qquad (21)$$

where $R^{[m]}(k) = V^{[m]T}(k) V^{[m]}(k)$.

Thus, an optimal union of all neurons' pool outputs of each cascade can be organized. It is clear that not only the multidimensional neo-fuzzy neurons can be used as such neurons, but also any other structures that implement the nonlinear mapping $R^{n+(m-1)g} \to R^g$.

## V. Experiment and Analysis

To demonstrate the efficiency of the proposed adaptive neuro-fuzzy system and its learning procedures (19) and (21), we have implemented a simulation test based on solving the chaotic Lorenz attractor identification. The Lorenz attractor is a fractal structure corresponding to the long-term behavior of Lorenz oscillator. The Lorenz oscillator is a 3-dimensional dynamical system that exhibits chaotic flow, well-known for its lemniscate shape. The map shows how a dynamical system state (three variables of a three-dimensional system) evolves over time in a complex non-repeating pattern.

The Lorenz attractor is described by the differential equation in the form

$$\begin{cases} \dot{x} = \sigma(y-x), \\ \dot{y} = x(r-z)-y, \\ \dot{z} = xy - bz. \end{cases} \quad (22)$$

We can rewrite (22) in the recurrent form:

$$\begin{cases} x(i+1) = x(i) + \sigma(y(i)-x(i))dt, \\ y(i+1) = y(i) + (rx(i)-x(i)z(i)-y(i))dt, \\ z(i+1) = z(i) + (x(i)y(i)-bz(i))dt \end{cases} \quad (23)$$

where parameter values are: $\sigma = 10, r = 28, b = 8/3, dt = 0.001$.

The data sample was obtained using (23) which consists of 10000 samples, where 7000 samples form a training set, 3000 samples form a checking set.

Fig.1 and Fig.2 present a time series output and a prediction value for 500 and 10000 iterations respectively (a dark color stays for a time series value, a light color stays for a prediction value).

Symmetric mean absolute percentage error (SMAPE) and mean square error (MSE), used for results evaluation, are shown in Fig.3 and Fig.4.

Fig.1 presents a time series output and a prediction value (a dark color stays for a time series value, a light color stays for a prediction value).

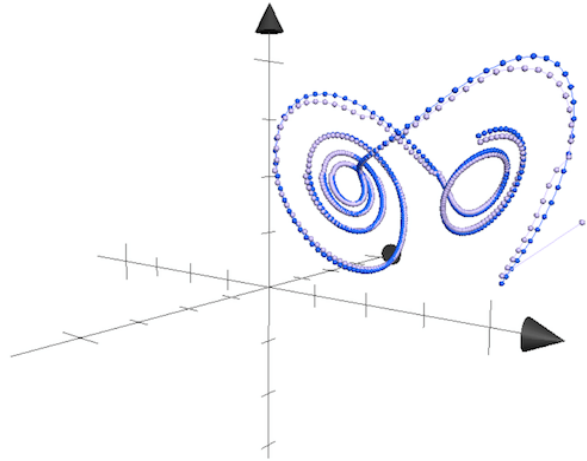

Fig.1. A visual representation of the predicted values (500 iterations)

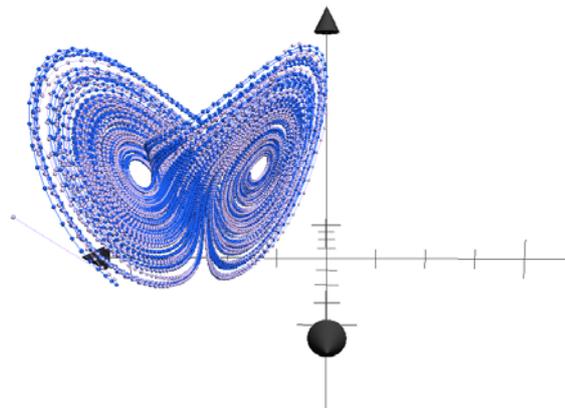

Fig.2. A visual representation of the predicted values (10000 iterations)

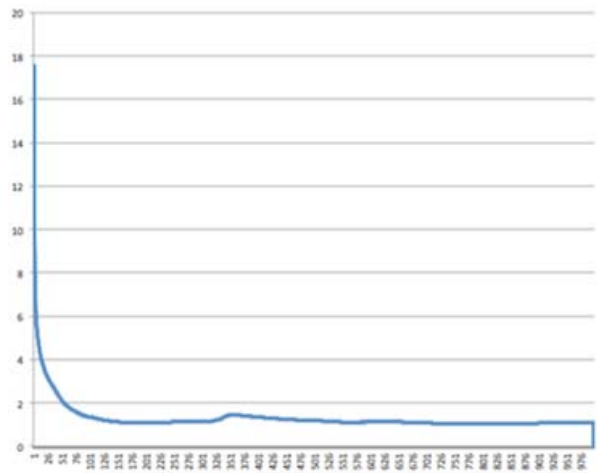
Fig.3. Symmetric mean absolute percentage error (SMAPE)

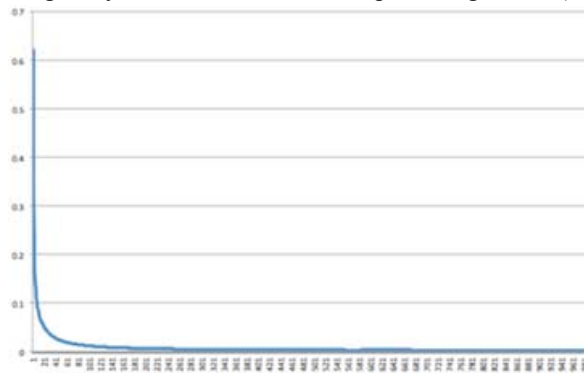
Fig.4. Mean squared error (SME)

By the $10000^{th}$ iteration the evolving system consists of 8-9 cascades and, as one can see in Fig.3 and Fig.4, the error curve is almost parallel to the Y-axis, which means the highest possible prediction accuracy, conditioned by the form of membership functions, is reached.

## VI. Conclusion

The new cascade system type of computational intelligence is proposed in the paper, in which nodes are multi-dimensional neo-fuzzy neurons that implement the multidimensional Takagi-Sugeno-Kang fuzzy reasoning. The adaptive learning algorithm for a multidimensional neo-fuzzy neuron is proposed that has both tracking and filtering properties. The distinguishing feature of the proposed system is that each cascade is formed by a set of neurons, and their outputs are combined with an optimization procedure of a special type. Thus, each cascade generates an optimal accuracy output signal. The proposed system which is essentially an evolving system of computational intelligence, makes it possible to process incoming data in an online mode unlike other traditional systems. This feature allows to process multidimensional non-stationary stochastic and chaotic signals under a priori and current uncertainty conditions as well as to provide the optimal forecast accuracy.

## References

[1] A. Cichocki, R. Unbehauen, *Neural Networks for Optimization and Signal Processing*. Stuttgart: Teubner, 1993.
[2] S. Haykin, *Neural Networks: A Comprehensive Foundation*. Upper Saddle River, New Jersey: Prentice Hall, 1999.
[3] N. Kasabov, *Evolving Connectionist Systems*. London: Springer-Verlag, 2003.
[4] E. Lughofer, *Evolving Fuzzy Systems – Methodologies, Advanced Concepts and Applications*. Berlin-Heidelberg: Springer-Verlag, 2011.
[5] S.E. Fahlman and C. Lebiere, "The cascade-correlation learning architecture," in *Advances Neural Information Processing Systems*, San Mateo, CA: Morgan Kaufman, 1990, pp. 524-532.
[6] L. Prechelt, "Investigation of the CasCor family of learning algorithms," in *Neural Networks*, vol. 10, 1997, pp. 885-896.
[7] R.J. Schalkoff, *Artificial Neural Networks*, New York: The McGraw-Hill Comp., 1997.
[8] E.D. Avedjan, G.V. Barkan and I.K. Levin, "Cascade neural networks," in *J. Avtomatika i Telemekhanika*, vol. 3, 1999, pp. 38-55.
[9] Ye. Bodyanskiy, A. Dolotov, I. Pliss and Ye. Viktorov, "The cascaded orthogonal neural network," in *Int. J. Information Science and Computing*, vol. 2, Sofia: FOI ITHEA, 2008, pp. 13-20.
[10] Ye. Bodyanskiy and Ye. Viktorov, "The cascaded neo-fuzzy architecture and its on-line learning algorithm," in *Int. J. Intelligent Processing*, vol. 9, Sofia: FOI ITHEA, 2009, pp. 110-116.

[11] Ye. Bodyanskiy and Ye. Viktorov, "The cascaded neo-fuzzy architecture using cubic-spline activation functions," in *Int. J. Information Theories and Applications*, vol. 16, no. 3, 2009, pp. 245-259.
[12] Ye. Bodyanskiy, Ye. Viktorov and I. Pliss, "The cascade growing neural network using quadratic neurons and its learning algorithms for on-line information processing," in *Int. J. Intelligent Information and Engineering Systems*, vol. 13, Rzeszov-Sofia: FOI ITHEA, 2009, pp. 27-34.
[13] V. Kolodyazhniy and Ye. Bodyanskiy, "Cascaded multi-resolution spline-based fuzzy neural network," *Proc. Int. Symp. on Evolving Intelligent Systems*, pp. 26-29, 2010.
[14] Ye. Bodyanskiy, O. Vynokurova and N. Teslenko, "Cascaded GMDH-wavelet-neuro-fuzzy network," *Proc $4^{th}$ Int. Workshop on Inductive Modelling*, pp. 22-30, 2011.
[15] Ye. Bodyanskiy, O. Kharchenko and O. Vynokurova, "Hybrid cascaded neural network based on wavelet-neuron," in *Int. J. Information Theories and Applications*, vol. 18, no. 4, 2011, pp. 35-343.
[16] Ye. Bodyanskiy, P. Grimm and N. Teslenko, "Evolving cascaded neural network based on multidimensional Epanechnikov's kernels and its learning algorithm," in *Int. J. Information Technologies and Knowledge*, vol. 5, no. 1, 2011, pp. 25-30.
[17] J-S.R. Jang, C.T. Sun and E. Mizutani, *Neuro-Fuzzy and Soft Computing: A Computational Approach to Learning and Machine Intelligence*, New Jersey: Prentice Hall, 1997.
[18] S. Wadhawan, G. Goel and S. Kaushik, "Data Driven Fuzzy Modelling for Sugeno and Mamdani Type Fuzzy Model using Memetic Algorithm," in *Int. J. Information Technology and Computer Science*, vol. 5, no. 8, 2013, pp. 24-37, doi: 10.5815/ijitcs.2013.08.03.
[19] T. Yamakawa, E. Uchino, T. Miki and H. Kusanagi, "A neo fuzzy neuron and its applications to system identification and prediction of the system behavior," *Proc. 2nd Int. Conf. on Fuzzy Logic and Neural Networks*, pp. 477-483, 1992.
[20] E. Uchino and T. Yamakawa, "Soft computing based signal prediction, restoration and filtering," in *Intelligent Hybrid Systems: Fuzzy Logic, Neural Networks and Genetic Algorithms*, Boston: Kluwer Academic Publisher, 1997, pp. 331-349.
[21] T. Miki and T. Yamakawa, "Analog implementation of neo-fuzzy neuron and its on-board learning," in *Computational Intelligence and Applications*, Piraeus: WSES Press, 1999, pp. 144-149.
[22] M. Barman and J.P. Chaudhury, "A Framework for Selection of Membership Function Using Fuzzy Rule Base System for the Diagnosis of Heart Disease," in *Int. J. Information Technology and Computer Science*, vol. 5, no.11, 2013, pp. 62-70, doi: 10.5815/ijitcs.2013.11.07.
[23] B. Widrow and Jr.M.E. Hoff, "Adaptive switching circuits," *URE WESCON Convention Record*, vol. 4, pp. 96-104, 1960.
[24] S. Kaczmarz, "Approximate solution of systems of linear equations," in *Int. J. Control*, vol. 53, 1993, pp. 1269-1271.
[25] L. Ljung, *System Identification: Theory for the User*, New York: Prentice-Hall, 1999.
[26] Ye. Bodyanskiy, I. Kokshenev and V. Kolodyazhniy, "An adaptive learning algorithm for a neo-fuzzy neuron," *Proc. 3rd Int. Conf. of European Union Soc. for Fuzzy Logic and Technology*, pp. 375-379, 2003.
[27] Ye. Bodyanskiy, O. Tyshchenko and D. Kopaliani, "Multidimensional non-stationary time-series prediction with the help of an adaptive neo-fuzzy model," in *Visnyk Nacionalnogo universytety "Lvivska politehnika"*, vol. 744, 2012, pp. 312-320.
[28] W.M. Caminhas, S.R. Silva, B. Rodrigues and R.P. Landim, "A neo-fuzzy-neuron with real time training applied to flux observer for an induction motor," *Proceedings 5th Brazilian Symposium on Neural Networks*, pp. 67-72, 1998.
[29] Ye.V. Bodyanskiy, I.P. Pliss and T.V. Solovyova, "Multistep optimal predictors of multidimensional non-stationary stochastic processes," in *Doklady AN USSR*, vol. A(12), 1986, pp. 47-49.
[30] Ye.V. Bodyanskiy, I.P. Pliss and T.V. Solovyova, "Adaptive generalized forecasting of multidimensional stochastic sequences," in *Doklady AN USSR*, vol. A(9), 1989, pp.73-75.
[31] Ye. Bodyanskiy and I. Pliss, "Adaptive generalized forecasting of multivariate stochastic signals," *Proc. Latvian Sign. Proc. Int. Conf.*, vol. 2, pp. 80-83, 1990.